\documentclass[10pt,twocolumn,letterpaper]{article}
\usepackage{cvpr}              
\usepackage{colortbl}
\usepackage{makecell}

\usepackage[dvipsnames]{xcolor}

\definecolor{cvprblue}{rgb}{0.21,0.49,0.74}
\usepackage[pagebackref,breaklinks,colorlinks,citecolor=cvprblue]{hyperref}
\usepackage{listings}

\def\paperID{4847} 
\def\confName{CVPR}
\def\confYear{2024}

\definecolor{codegreen}{rgb}{0,0.5,0}
\definecolor{codeblue}{rgb}{0.25,0.5,0.5}
\definecolor{codegray}{rgb}{0.6,0.6,0.6}

\title{LORS: Low-rank Residual Structure for Parameter-Efficient Network Stacking}

\author{Jialin Li,~~~Qiang Nie,~~~Weifu Fu,~~~Yuhuan Lin,~~~Guangpin Tao,~~~Yong Liu,~~~Chengjie Wang\\
Youtu Lab, Tencent\\
{\tt\small $\{$jarenli, ryanwfu, gleelin, guangpintao, choasliu, jasoncjwang$\}$@tencent.com, qnie.cuhk@gmail.com}
}

\begin{document}
\maketitle
\begin{abstract}
Deep learning models, particularly those based on transformers, often employ numerous stacked structures, which possess identical architectures and perform similar functions. While effective, this stacking paradigm leads to a substantial increase in the number of parameters, posing challenges for practical applications. In today's landscape of increasingly large models, stacking depth can even reach dozens, further exacerbating this issue. To mitigate this problem, we introduce \textbf{LORS} (\textbf{LO}w-rank \textbf{R}esidual \textbf{S}tructure). LORS allows stacked modules to share the majority of parameters, requiring a much smaller number of unique ones per module to match or even surpass the performance of using entirely distinct ones, thereby significantly reducing parameter usage. We validate our method by applying it to the stacked decoders of a query-based object detector, and conduct extensive experiments on the widely used MS COCO dataset. Experimental results demonstrate the effectiveness of our method, as even with a 70\% reduction in the parameters of the decoder, our method still enables the model to achieve comparable or even better performance than its original. 
\end{abstract}    
\section{Introduction}

\label{sec:intro}
\begin{figure}[t]
  \centering
   \includegraphics[width=0.8\linewidth]{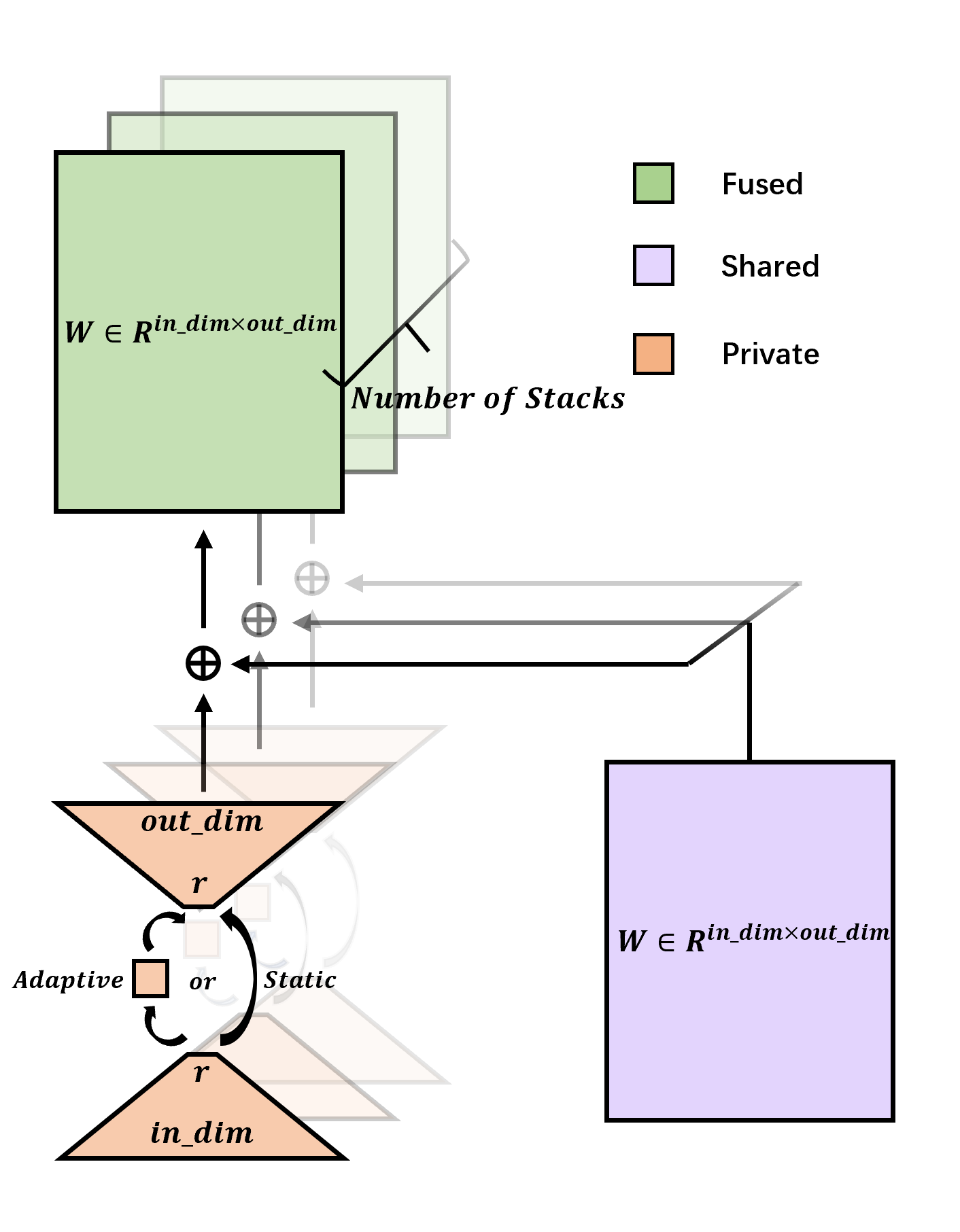}
   \caption{The LORS calculation process, which could be adaptive or static, depending on whether an adaptively generated kernel is used in the matrix manipulation for private parameters. }
   \label{fig:LORA}
\end{figure}

In the current era of prosperity for large models, a common issue is the significant increase in the number of parameters, which presents challenges for training, inference, and deployment. Various methods have been proposed to reduce the number of parameters in models, such as knowledge distillation~\cite{hinton2015distilling,gou2021knowledge}, which compresses large models into smaller ones while trying to preserve their performance but may still lead to a decrease in model capacity; pruning~\cite{han2015deep, zhu2017prune}, which removes redundant parameters from the model but can affect the model's stability; quantization~\cite{courbariaux2015binaryconnect}, which reduces the numerical precision of model parameters to lower storage and computation but may cause model accuracy loss; and parameter sharing~\cite{lan2019albert}, which reduces the number of parameters by sharing them across different layers but may limit the model's expressiveness.

Different from the aforementioned methods, we have observed an important fact contributing to the large number of parameters: the widespread use of stacking in neural networks. Stacking refers to those modules that have identical architectures and perform the same or similar functions, but possess different parameters due to random initialization as well as training updates. Examples of stacking can be found in numerous prominent neural networks, like the classic ResNet model~\cite{he2016deep} and Transformers~\cite{vaswani2017attention}. Particularly, Transformers heavily rely on stacked structures and typically employ completely identical multi-layer stacks in both encoders and decoders. It now serves as an indispensable component of many excellent models in fields such as computer vision and natural language processing.

Although stacking is powerful for enhancing model capacity, as demonstrated by large language models, it also naturally leads to a sharp increase in the number of parameters. For example, GPT-3~\cite{brown2020language} utilizes 175 billion parameters and consists of 96 layers of stacked Transformer~\cite{vaswani2017attention} layers. How can we enjoy the benefits of stacking while reducing the required number of parameters? We notice that stacked decoders have identical structures and similar functions, indicating that there should be some commonality among their parameters. However, since they handle different input and output distributions, there must also be unique aspects to their parameters. Therefore, a natural idea is: it may be possible to represent the shared aspects with shared parameters while allowing each stacked module to retain only the parameters that capture its unique characteristics, thereby reducing the overall parameter usage.

Based on the above considerations, we propose to decompose the parameters of stacked modules into two parts: shared ones representing the commonality and private ones capturing the specific characteristics. Shared parameters are available for all modules and trained jointly by them, while the private parameters are separately owned by each module. We suppose that the way of sharing parameters related to the commonality may reduce the number of parameters while maintaining the performance of the model. To achieve this goal, we introduce the concept of Low-rank Residual Structure (LORS), inspired by the approach of LoRA~\cite{hu2021lora}. LORS essentially adds the unique parameters to the shared ones, just like residual connections add residual information to the features. While LoRA is originally designed for fine-tuning, we train our LoRA-like operation on parameters from scratch.
This approach allows us to effectively reduce the overall parameter usage while maintaining the performance of the model, paving the way for more parameter-efficient network stacking.

To validate our idea, we choose AdaMixer~\cite{gao2022adamixer}, a strong query-based object detector, as our experimental subject. It contains a large number of both adaptive and static parameters in its stacked decoders, making it an ideal candidate to demonstrate LORS's effectiveness. The difference between adaptive and static parameters lies in whether they change with different inputs. Our goal is to show that LORS can effectively reduce the overall usage of both types of parameters while maintaining the model's performance. We conducted extensive experiments on this detector, illustrating that our method succeeded in reducing up to 70\% of parameters in AdaMixer's decoders, while still enabling the model to achieve comparable or even superior performance compared to its vanilla version.
In summary, our contributions can be concluded as :
\begin{itemize}
\item We propose a novel low-rank residual structure, named LORS, for network stacking, which can greatly reduce the number of parameters while maintaining or even improving the performance compared to the vanilla one.
\item We introduce effective methodologies for reducing both  static and adaptive generated parameters in stacked structures, which makes our proposed LORS a more versatile method. 
\item Our method holds the potential to serve as one of the basic network structures for large models that are affected by the issue of excessive parameters due to stacking, rendering them more parameter-efficient, thereby facilitating an easier implementation in practical applications.
\end{itemize}
\section{Related Work}
\label{sec:2_related_work}

\noindent\textbf{Models with stacked structures.} Many nerual networks partially or extensively employ stacked modules. CNN-based models~\cite{sultana2020review}, for instance, are widely applied across various computer vision tasks, such as classification~\cite{simonyan2014very,huang2017densely,xie2017aggregated,he2016deep}, detection~\cite{redmon2016you,redmon2017yolo9000,ren2015faster,he2017mask,tian2019fcos, law2018cornernet}, segmentation~\cite{zhao2017pyramid,chen2018encoder,long2015fully,ge2023beyond}, etc. These models often use stacked smaller modules within larger components. 
Another kind of models primarily utilize Multi-Layer Perceptrons (MLPs), such as the MLP-Mixer~\cite{tolstikhin2021mlp}, whose main body comprises dozens of completely identical stacked mixer layers. 
Moreover, since its invention, Transformer~\cite{vaswani2017attention} has been successfully applied to diverse domains, including computer vision~\cite{carion2020end, zhu2020deformable,liu2022dab,sun2021sparse}, natural language processing~\cite{brown2020language,yang2019xlnet,devlin2018bert}, multi-modal learning~\cite{wang2023visionllm,chen2023shikra,radford2021learning,li2022blip}, etc. In the thriving arena of large language models, Transformer-based structures are indeed indispensable for many cutting-edge works~\cite{devlin2018bert,radford2019language,lewis2019bart,workshop2022bloom,thoppilan2022lamda}. Transformers are usually used to form stacked multi-layers encoders or decoders, within each layer being structurally identical.

\noindent\textbf{LoRA and Its Variants.} 
LoRA~\cite{hu2021lora} is a technique proposed for fine-tuning large language models, it introduces low-rank decomposition matrices in the Transformer~\cite{vaswani2017attention} architecture, significantly reducing trainable parameters and GPU memory requirements. 
Afterward, a series of works with modifications as well as improvement have been proposed.
AdaLoRA~\cite{zhang2023adaptive} adaptively allocates parameter budgets among weight matrices based on their importance scores and uses singular value decomposition for incremental updates, leading to an improved fine-tuning performance, especially in low budget settings.
DyLoRA~\cite{valipour2023dylora} introduces a dynamic low-rank adaptation technique that trains LoRA blocks for a range of ranks instead of a single rank, addressing the issues of fixed block size and rank optimization in LoRA. 
Delta-LoRA~\cite{zi2023deltalora} presents a parameter-efficient approach to fine-tune large language models by updating not only the low-rank matrices but also the pre-trained weights using the delta of the product of two low-rank matrices. 
LoRA-FA~\cite{zhang2023lorafa} introduces a memory-efficient fine-tuning method that reduces activation memory without performance degradation or expensive recomputation, it achieves this by freezing the projection-down weight and updating the projection-up weight in each LoRA layer.
ReLoRA~\cite{lialin2023stack} is a parameter-efficient method for training large neural networks using low-rank updates.
GLoRA~\cite{chavan2023oneforall}, or Generalized LoRA, is an advanced approach for universal parameter-efficient fine-tuning tasks, which enhances the low rank adaptation technique by employing a generalized prompt module to optimize pre-trained model weights and adjust intermediate activations.
VeRA~\cite{kopiczko2023vera}, or Vector-based Random Matrix Adaptation, further compresses the training parameters compared to the vanilla LoRA~\cite{hu2021lora}. It achieves this by utilizing shared low-rank matrices across all layers and learning small scaling vectors.

Inspired by the aforementioned LoRA series works, we propose our method called LORS (Low Rank Residual Structure) for parameter efficiency. However, our approach is fundamentally distinct from the LoRA series works. While LoRA uses different low-rank weights to fine-tune different pretrained weights, LORS adds different low-rank weights per layer to the same common weights shared by all layers, thus creating parameter-efficient stacked models trained from scratch, requiring no pretrained weights. 
\section{Approach}

In this section, we initially revisit the foundational works upon which our methodology and experiments are primarily based, encompassing the LoRA mechanism~\cite{hu2021lora}, query-based object detection~\cite{carion2020end}, and the structure of AdaMixer's decoders~\cite{gao2022adamixer}. Then we mathematically formulate our proposed LORS, which comprises two parts: the adaptive one and the static one. Subsequently, we elaborate on how the LORS method is applied to reduce the parameters of AdaMixer's decoders. Lastly, we provide a quantitative analysis of the parameter reduction percentage attained by our proposed method.

\subsection{Preliminary}

\noindent\textbf{The mechanism of LoRA. } The Low-Rank Adaptation (LoRA)~\cite{hu2021lora} technique is a novel approach designed to adapt large pre-trained language models for specific tasks. The key idea of LoRA is to introduce a low-rank parameter matrix which is able to captures task-specific knowledge while the original pre-trained parameters remain fixed. 

Mathematically, given a pre-trained parameter matrix $W \in \mathbb{R}^{d \times h}$, LoRA uses a low-rank matrix $B \in \mathbb{R}^{d \times r}$ and a projection matrix $A \in \mathbb{R}^{r \times h}$ to adapt $W$, where $r \ll d, h$. The adapted parameter matrix is then given by 
\begin{equation}
W + \Delta{W} = W + BA
\end{equation}
where $BA$ captures the task-specific knowledge.

The key advantage of LoRA is that it can significantly reduce the number of parameters that need to be fine-tuned, thereby reducing the computational cost and lowering down the memory requirement. In some cases, even single-digit values of $r$ are sufficient to fine-tune the model to the desired state, which is often tens of times less expensive than training the parameters in $W$. Furthermore, by keeping the original parameters fixed, LoRA avoids catastrophic forgetting, a common issue in fine-tuning large models.

\noindent\textbf{Query-based object detection.} 
In the realm of object detection, query-based detectors have established a new paradigm~\cite{gao2021fast,zhang2022dino,sun2021sparse,liu2022dab,zhu2020deformable,gao2022adamixer}. Unlike traditional detectors which rely on anchor boxes or sliding windows, query-based models utilize a set of learnable queries to interact with image feature maps. This interaction can be formalized with attention~\cite{vaswani2017attention} operations as
\begin{equation}
    Q_{\text{updated}} = \text{Attention}(Q, K(V))
\end{equation}
where $Q$, $K$, and $V$ represent the queries, keys, and values. The learnable queries $Q$ are used to predict object classes and bounding boxes ultimately, while $K$ and $V$ typically originate from encoded image features. It is a common practice to continuously refine $Q$ through interactions with $K$ and $V$ using successive decoding layers. These layers are usually composed of structurally identical decoders.

\noindent\textbf{Decoders of AdaMixer.} \label{ada_decoder}
AdaMixer~\cite{gao2022adamixer} is a query-based detector that features an innovative decoder design. This design includes Adaptive Channel Mixing (ACM) and Adaptive Spatial Mixing (ASM) methods, which greatly enhance its performance~\cite{gao2022adamixer}.

Given a sampled feature $\mathbf{x}\in\mathbb{R}^{P_{\rm in}\times C}$, where $C=d_{\rm feat}/g$, and $g$ denotes the number of sampling groups. This sampled feature is obtained through an operation called group sampling. This operation divides the feature space channel $d_{\rm feat}$ into $g$ groups and performs individual sampling for each group. Then, ACM (Adaptive Channel Mixing) utilizes a weight adapted by the object query $\mathbf{q}$ to transform feature $\mathbf{x}$ in the channel dimension, enhancing channel semantics~\cite{gao2022adamixer}:
\begin{align}
    M_c &= {\rm Linear}(\mathbf{q}) \in \mathbb{R}^{C\times C} \\
    {\rm ACM}(\mathbf{x})&= {\rm ReLU}({\rm LayerNorm}(\mathbf{x}M_c))
\end{align}
where LayerNorm stands for Layer Normalization~\cite{ba2016layer}.

Next is the ASM (Adaptive Spatial Mixing) process, which aims to enable the adaptability of the object query $\mathbf{q}$ to spatial structures of sampled features~\cite{gao2022adamixer} by applying adaptive transformation to the spatial dimension: 
\begin{align}
    M_s &= {\rm Linear}(\mathbf{q}) \in \mathbb{R}^{P_{\rm in}\times P_{\rm out}} \\
    {\rm ASM}(\mathbf{x}) &= {\rm ReLU}({\rm LayerNorm}(\mathbf{x}^T M_s)),
\end{align}
Both ACM and ASM train independent parameters for each sampling group, and finally the output with the shape $\mathbb{R}^{g\times C\times P_{\rm out}}$ is flattened and transformed to the $d_{q}$ dimension by a linear layer $L_{\text{output}}$ to add back to the object query. 

ACM, ASM and the output linear transformation $L_{output}$ possess significantly more parameters compared to the decoder's other operations, making them the main contributors to the number of parameters. Hence, we choose them as the target components for validating the effectiveness of our LORS method in parameter reduction.
\lstset{
  backgroundcolor=\color{white},
  basicstyle=\fontsize{7.5pt}{8.5pt}\fontfamily{lmtt}\selectfont,
  columns=fullflexible,
  breaklines=true,
  captionpos=b,
  commentstyle=\fontsize{8pt}{9pt}\color{codegray},
  keywordstyle=\fontsize{8pt}{9pt}\color{codegreen},
  stringstyle=\fontsize{8pt}{9pt}\color{codeblue},
  frame=tb,
  otherkeywords = {self},
}

\begin{figure}[ht]
\begin{lstlisting}[language=python]
def non_dynamic_fused_weight(W_shared, A, B):
    # W_shared: (1, d, h)
    # A: (G, r, h) 
    # B: (G, d, r)

    # private weight: (G, d, h)
    W_private = B @ A

    # sum at group dimension to get fused weight: (d, h)
    W = concat([W_shared, W_private], dim=0).sum(axis=0)

    return W
\end{lstlisting}
\vspace{-1em}
\caption{Pseudo-code for obtaining a static weight parameter for one layer. }
\label{fig:non-dynamic}
\vspace{3mm}
\end{figure}

\subsection{Formulation of Our Method}
\label{formulation}
The complete computational process of LORS is illustrated in Figure \ref{fig:LORA}. Notably, LORS computation is divided into two types: adaptive and static. In the context of AdaMixer, the term "adaptive" indicates whether the transformation matrix depends on the object query. After completing the formalization of LORS here, we will further elaborate on how it is applied to the tasks in our setting.

\noindent\textbf{Static Low Rank Residual Structure (LORS\textsuperscript{T})}. Suppose there are $N$ stacked layers of modules with identical architecture, and $W_i \in \mathbb{R}^{d \times h}$ be a parameter matrix belonging to the $i$-th layer. Then, we have:
\begin{equation}
    W_i = W^{\text{shared}} + W^{\text{private}}_i
\end{equation}
Here, $W^{\text{shared}} \in \mathbb{R}^{d \times h}$ represents the shared parameters across all stacked layers, while $W^{\text{private}}_i$ denotes the layer-specific parameters for the $i$-th layer, which is calculated as follows:
\begin{equation}
    W^{\text{private}}_i = \sum_{k=1}^{K} B_{ik} A_{ik}
\end{equation}
$B_{ik} \in \mathbb{R}^{d \times r} $, $A_{ik} \in \mathbb{R}^{r \times h}$, and $r \ll d, h$. $K$ represents the number of parameter groups used to compute $W^{\text{private}}_i$. Pseudo-code of LORS\textsuperscript{T} computing $W^{\text{private}}_i$ for a specific layer $i$ can be seen in Figure \ref{fig:non-dynamic}.

\begin{figure}[ht]
\begin{lstlisting}[language=python]
def dynamic_fused_weight(q, A, B):
    # q: (N, C)
    # A: (1, G, r, h) 
    # B: (1, G, d, r)

    # shared weight: (N, 1, d, h)
    W_shared = linear1(pro_feats).reshape(N, 1, d, h)
    # E: (N, G, r, r)
    E = linear2(pro_feats).reshape(N, G, r, r)
    # private weight: (N, G, d, h)
    W_private = B @ E @ A

    # sum at group dimension to get fused weight: (N, d, h)
    W = concat([W_shared, W_private], dim=1).sum(axis=1)

    return W
\end{lstlisting}
\vspace{-1em}
\caption{Pseudo-code for obtaining an adaptive weight parameter for one layer. }
\label{fig:dynamic}
\vspace{3mm}
\end{figure}

\noindent\textbf{Adaptive Low Rank Residual Structure (LORS\textsuperscript{A})}. 
Let $\hat{W}_i \in \mathbb{R}^{d \times h}$ be an adaptive generated parameter in $i$-th stacked layer, it is similarly calculated as :
\begin{equation}
    \hat{W}_i = \hat{W}^{\text{shared}} + \hat{W}^{\text{private}}_i
\end{equation}
where the cross-layer shared parameter $\hat{W}^{\text{shared}} \in \mathbb{R}^{d \times h}$ and the layer-specific parameter $\hat{W}^{\text{private}}_i \in \mathbb{R}^{d \times h}$ for each layer are both calculated based on $q$:
\begin{align}
    \hat{W}^{\text{shared}} &= {\rm Linear}(\mathbf{q}) \in \mathbb{R}^{d\times h} \\
    \hat{W}^{\text{private}}_i &= \sum_{k=1}^{K} \hat{B}_{ik} \hat{E}_{ik} \hat{A}_{ik} \label{formula:adpt_private}\\ 
    \hat{E}_{ik} &= {\rm Linear}(\mathbf{q}) \in \mathbb{R}^{r\times r} 
\end{align}
where $\hat{B}_{ik} \in \mathbb{R}^{d \times r} $ and $\hat{A}_{ik} \in \mathbb{R}^{r \times h}$, $r \ll d, h$. Pseudo-code of LORS\textsuperscript{A} calculating $\hat{W}^{\text{private}}_i$ for a specific layer $i$ can be seen in Figure \ref{fig:dynamic}.

\subsection{Applying LORS to AdaMixer's Decoders}

We apply LORS to the parameters of linear transformations belonging to ACM, ASM, and $L_{\text{output}}$ in each of AdaMixer's decoders. These parameters are explained in \ref{ada_decoder}.

\begin{figure}[t]
  \centering
   \includegraphics[width=1.0\linewidth]{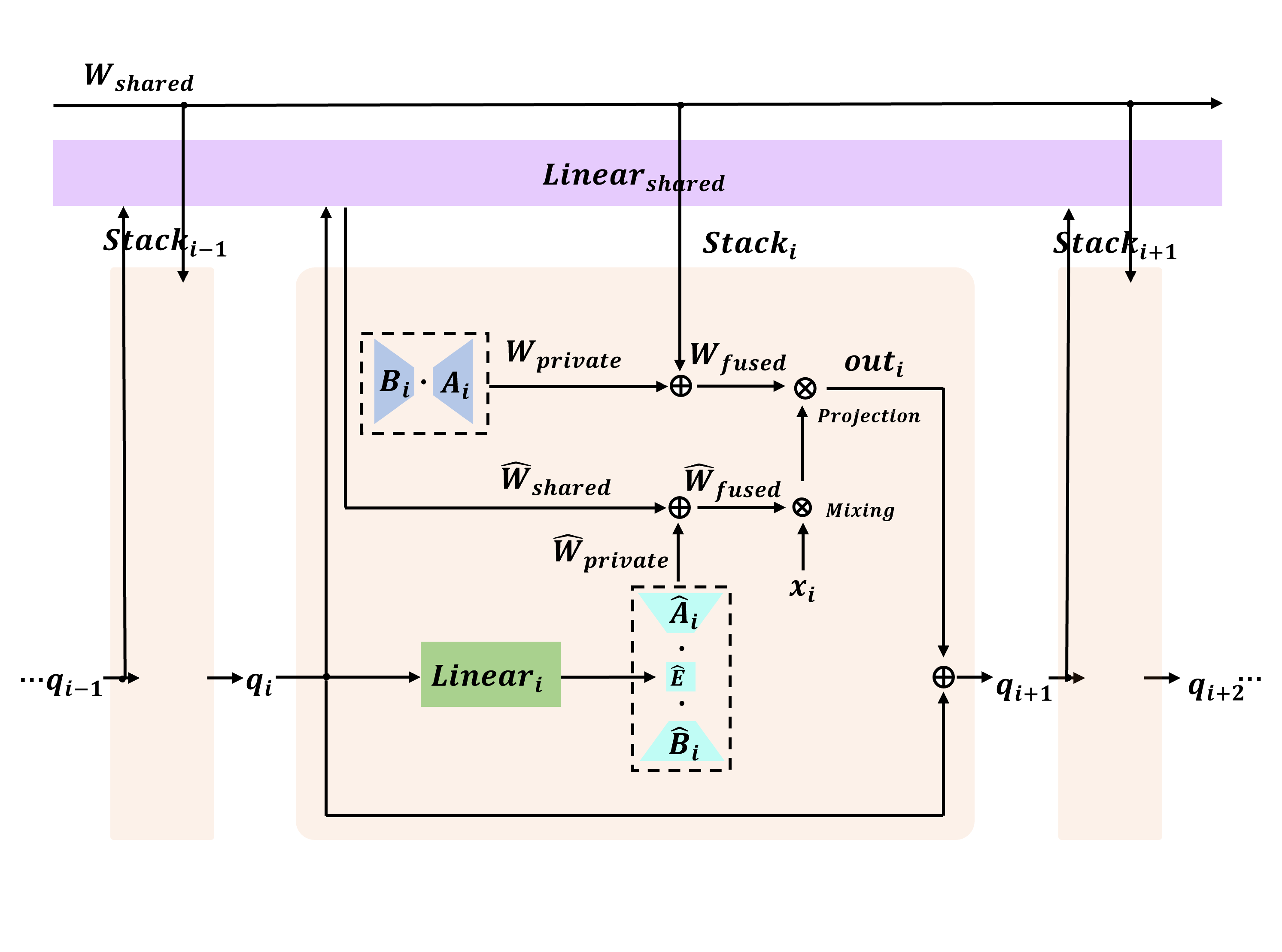}
   \caption{The overall pipeline of our proposed LORS, consisting of both adaptive and static parts, each further composed of shared and private components, works collaboratively. The figure illustrates the entire computation process within one layer of the stacked layers, with an enlarged example of the $i$-th layer for demonstration.}
   \label{fig:LORS_PPL}
\end{figure}

The overall pipeline of LORS running in AdaMixer is illustrated in Figure \ref{fig:LORS_PPL}. Specifically, for each group of sampling points, LORS\textsuperscript{A} (see \ref{formulation}) is used to reduce parameters in $M_c$ (from $\mathbb{R}^{d_q}$ to $\mathbb{R}^{C\times C}$) of ACM and $M_s$ of ASM (from $\mathbb{R}^{d_q}$ to $\mathbb{R}^{P_{\rm in}\times P_{\rm out}}$), and LORS\textsuperscript{T}(see \ref{formulation}) to minimize parameters in $L_{\text{output}}$ (from $\mathbb{R}^{C\times P_{\rm out}}$ to $\mathbb{R}^{d_q}$).

Thus, the parameter quantities of $M_c$, $M_s$, and $L_{\text{output}}$ are 
$d_q \times C \times C$, 
$d_q \times P_{\rm in} \times P_{\rm out}$ and 
$d_q \times C \times P_{\rm out} $ respectively.
When the sampling strategy consists of 2 groups with 64 points each, which is the default setting in our experiments, the values of the variables are as follows: $d_q=256$, $C=64$, $P_{\rm in}=64$, and $P_{\rm out}=128$. It can be easily calculated that the number of parameters for each of $M_c$, $M_s$, and $L_{\text{output}}$ exceeds one million. 
Indeed, these three components collectively account for the most of the total parameter in the AdaMixer model with ResNet-50 as the backbone, whereas they are also the primary drivers of enhanced model performance. This is what motivates us to conduct LORS experiments on them.
The effects of LORS applied to them  will be discussed in detail in \ref{analysis_param}.

\subsection{Analysis on Parameter Reduction}
\label{analysis_param}
Let $W \in \mathbb{R}^{d \times h}$ be a weight parameter that exists in every layer of stacked structures, $N$ is the number of stacked layers. If static, it originally has $d \times h$ parameters, while using LORS\textsuperscript{T} requires $\frac{1}{N} \times d \times h + K \times(d \times r + r \times h)$ parameters  on average per layer; if it is adaptive, generating it by $q$ with linear transformation requires $d_q \times d \times h$ parameters, where $d_q$ is the dimension of $q$, and using LORS\textsuperscript{A} requires $ \frac{1}{N} \times d_q \times d \times h + K \times(d_q \times r^2 + d \times r + r \times h)$ parameters  on average per layer. To more intuitively display the parameter reduction effect of LORS, we set $d_q=256$, $d=64$, $h=128$, $K=2$ for the ASM process and $d=2 \times 128 \times 128 $, $h=256$, $K=1$ for the $L_{output}$, which is actually the case of AdaMixer's decoders in our experiment, and show the parameter reduction for different $r$ values using LORS\textsuperscript{T} and LORS\textsuperscript{A} in Table \ref{tab:analysis}.

\begin{table}[ht]
  \centering
  \resizebox{0.95\columnwidth}{!}
  {
  \begin{tabular}{l c c c c}
    \toprule
     & r=4 & r=8 & r=16 & r=32\\
    \midrule
    W/ . LORS\textsuperscript{T} & 1.53M& \cellcolor{gray!20} 1.66M&  1.93M& 2.46M\\
    W/O. LORS\textsuperscript{T} & 8.39M& \cellcolor{gray!20} 8.39M&  8.39M& 8.39M\\
    Percentage & 18.3\% & \cellcolor{gray!20} 19.8\% &  23.0\% & 29.3\% \\
    \midrule
    W/ . LORS\textsuperscript{A} & 0.36M& 0.39M& \cellcolor{gray!20} 0.49M& 0.89M\\
    W/O. LORS\textsuperscript{A} & 2.10M& 2.10M& \cellcolor{gray!20} 2.10M& 2.10M\\
    Percentage & 17.2\% & 18.6\% & \cellcolor{gray!20} 23.3\% & 42.4\% \\
    \bottomrule
  \end{tabular}
  }
  \caption{Analysis of the parameter reduction effect of LORS\textsuperscript{T} and LORS\textsuperscript{A} with varying rank $r$ values. Our default selections are colored gray. Here the average amount of shared parameters over each layer has already been taken into account.}
  \label{tab:analysis}
\end{table}



\section{Experiments}
In this section, we first explain the implementation details of our experiments, including the dataset, training specifics, and different parameter initialization methods for static and adaptive LORS. Next, we present the main experimental results. Finally, we perform ablation studies to examine LORS's design and its individual components.
\label{sec:4_experiments}

\subsection{Implementation Details}

\noindent\textbf{Dataset.}
Our experiments were conducted on the widely-used MS COCO~\cite{lin2014microsoft} dataset in mmdetection codebase~\cite{chen2019mmdetection}, using the standard metrics for object detection. All models were trained on the \textit{train2017} split ($\sim$118k images) and then evaluated on the \textit{val2017} split (5k images).

\noindent\textbf{Training.}
We use the AdamW optimizer~\cite{loshchilov2017decoupled} with a weight decay of 0.0001 and train all models on 8 Nvidia V100 GPUs with a batch size of 16 and a learning rate of $2.5 \times e^{-5}$. 
Models are trained for either 12 or 36 epochs, with learning rate decreasing by a factor of 10 at epochs 8 and 11 for 12-epoch training, and at epochs 24 and 33 for 36-epoch training. 
The low-rank values are set as $r=16$ for LORS\textsuperscript{A} and $r=8$ for LORS\textsuperscript{T}.
The number of parameter groups is set as as $K=[1,1,2,2,3,3]$ for LORS\textsuperscript{A} in all experiments, applied to ACM and ASM in AdaMixer's decoders, and $K=[1,1,1,1,1,1]$ for $L_{output}$ in all experiments.
We divide the feature channels into 2 groups with 64 sampling points each instead of AdaMixer's default 4 groups with 32 sampling points each, aiming to increase the parameter compressible space for LORS, which does not improve performance according to both AdaMixer paper~\cite{gao2022adamixer} and our experiments. 
Backbones are initialized with ImageNet-1k~\cite{deng2009imagenet} pre-trained models, and LORS-related parameter initialization is detailed below,
remaining parameters are Xavier-initialized~\cite{glorot2010understanding}. Finally, all other aspects about model training, like the data augmentation, the loss function, etc., just follow AdaMixer's settings~\cite{gao2022adamixer}.

\noindent\textbf{Initialization Strategies}
\label{subsec:init_strategies.}
We tried various initialization methods for each component in LORS and determined the overall initialization method as follows:
\begin{itemize}
    \setlength\itemindent{1em}
    \item \textbf{LORS\textsuperscript{T}}: For static LORS, we employ Kaiming initialization~\cite{he2015delving} for $W^{shared}$ and each $B$, and zero initialization for each $A$.
    \item \textbf{LORS\textsuperscript{A}}: For adaptive LORS, we apply Kaiming initialization~\cite{he2015delving} to the linear transformation weights forming each $\hat{W}^{\text{shared}}$, as well as each $\hat{B}$ and $\hat{A}$. 
    Additionally, we use zero initialization for the linear transformation weights forming each $\hat{E}$.
\end{itemize}

\subsection{Main Results}

\begin{table*}[ht]
  \centering
  \resizebox{1.0\textwidth}{!}{
  \begin{tabular}{l|c|c|cc|ccc|ccccccc}
    \toprule
    Method & Queries & Epochs  & \begin{tabular}{@{}c@{}}Decoder\\ Params(M)\end{tabular} & Params(M) & \begin{tabular}{@{}c@{}}Decoder\\ GFLOPs\end{tabular} & GFLOPs & \begin{tabular}{@{}c@{}}Training\\ Hours\end{tabular} & $AP$ & $AP_{50}$ & $AP_{75}$ & $AP_{s}$ & $AP_{m}$ & $AP_{l}$ \\ 
    \midrule
    AdaMixer & 100 & 12  & 110 & 135 & \textbf{12} & \textbf{104} & \textbf{8.5h} &\textbf{42.7} & \textbf{61.5} & 45.9 & 24.7 & 45.4 & \textbf{59.2} \\
    AdaMixer + LORS & 100 & 12 & \textbf{35} & \textbf{60} & 18 & 110 & 9.5h &42.6 & 61.4 & \textbf{46.0} & \textbf{25.0} & \textbf{45.6} & 58.5 \\ 
    \midrule
    AdaMixer & 300 & 12 & 113 & 139 & \textbf{39} & \textbf{132} & \textbf{10h} & \textbf{44.1} & \textbf{63.4} & 47.4 & 27.0 & 46.9 & \textbf{59.5}\\ 
    AdaMixer + LORS & 300 & 12 & \textbf{35} & \textbf{60} & 56 & 149 & 12h & \textbf{44.1} & 63.0 & \textbf{47.7} & \textbf{27.8} & \textbf{47.0} & \textbf{59.5}\\ 
    \bottomrule
  \end{tabular}
  }
  \caption{\textbf{1× training scheme} performance on COCO 2017 val set with ResNet-50 as backbone. AdaMixer with LORS can achieve competitive results while employing a notably reduced number of parameters. FPS is obtained with a single Nvidia V100 GPU.}
  \label{table_main1}
\end{table*}

\begin{table*}[ht]
  \centering
  \resizebox{1.0\textwidth}{!}{
  \begin{tabular}{l|c|c|c|c|c|cccccc}
    \toprule
    Method & backbone & Queries & Epochs & Params(M)  & GFLOPs & $AP$ & $AP_{50}$ & $AP_{75}$ & $AP_{s}$ & $AP_{m}$ & $AP_{l}$ \\
    \midrule
    AdaMixer & R-50 & 100 & 36 & 135 & \textbf{104} & 43.2 & 61.8 & 46.7 &  25.0 & 46.1 &  58.8 \\
    AdaMixer + LORS & R-50 & 100 & 36 & \textbf{60} & 110 & \textbf{43.7} & \textbf{62.3} & \textbf{47.3} & \textbf{25.5} & \textbf{46.4} & \textbf{60.0}\\
    \midrule
    AdaMixer & R-50 & 300 & 36 & 139 & \textbf{132} & 47.0 & 66.0 & 51.1 & 30.1 & \textbf{50.2} & 61.8  \\
    AdaMixer + LORS & R-50 & 300 & 36 & \textbf{60} & 149 & \textbf{47.6} & \textbf{66.6} & \textbf{52.0} & \textbf{31.1} & \textbf{50.2} & \textbf{62.5} \\
    \midrule
    AdaMixer & R-101 & 300 & 36 & 158 & \textbf{208} & 48.0 & 67.0 & 52.4 & 30.0 & 51.2 & 63.7 \\
    AdaMixer + LORS & R-101 & 300 & 36 & \textbf{79} & 225 & \textbf{48.2} & \textbf{67.5} & \textbf{52.6} & \textbf{31.7} & \textbf{51.3} & \textbf{63.8}\\
    \midrule
    AdaMixer & Swin-S & 300 & 36 & 164 & \textbf{234} & 51.3 & 71.2 & 55.7 & 34.2 & 54.6 & 67.3  \\
    AdaMixer + LORS & Swin-S & 300 & 36 & \textbf{85} & 250 & \textbf{51.8} & \textbf{71.6} & \textbf{56.4} & \textbf{35.4} & \textbf{55.0} & \textbf{68.4} \\
    \bottomrule
  \end{tabular}
  }
  \caption{\textbf{3× training scheme} performance on COCO 2017 val set, considering different combinations of backbone and query numbers. Longer training time allows LORS to be fully trained and perform better. The comprehensive improvement in performance and the significant reduction in parameters demonstrate the effectiveness of LORS. The best results in each category are highlighted in bold.}
  \label{table_main2}
\end{table*}

To reliably demonstrate that LORS can reduce the number of parameters in stacked structures while maintaining model performance, which is measured by AP metric, we follow common practices in the object detection field and present experimental comparisons in two categories based on training schemes. The first involves training for 12 epochs using a weaker data augmentation, and the second involves training for 36 epochs using a stronger data augmentation, both of which are identical to those in AdaMixer~\cite{gao2022adamixer}.

Table \ref{table_main1} presents the comparison of performance with and without the LORS technique, under a 1× training scheme. Metrics such as the number of parameters, GFLOPs, and Average Precision (AP) at various scales are evaluated. 
When the AdaMixer model is combined with LORS, it demonstrates a remarkable reduction in the number of parameters. This reduction is consistently observed across various query quantities and different training epochs, encompassing both the decoder and overall model. For example, when trained with 100 queries for 12 epochs, the AdaMixer+LORS model uses only 35M decoder parameters and 60M total parameters, compared to 110M and 135M respectively in the AdaMixer model. This shows reduction of approximately 70\% in the decoder's parameters, which is substantial.
However, the AdaMixer+LORS model shows a slight increase in GFLOPs. 
In terms of model performance, the AdaMixer+LORS model achieves competitive results, even with a significantly reduced number of parameters. When trained with 100 queries, although the AP of 42.6 is slightly lower than the 42.7 AP of the vanilla AdaMixer, it still slightly outperforms in $AP_s$ and $AP_m$. When trained with 300 queries, the use of LORS does not affect the performance and allows an advantage in all $AP_s$, $AP_m$ and $AP_l$.
These findings suggest that even under conditions of limited model training and data augmentation, LORS does not impact training result, the model can still converge quickly to its original limit.
Additionally, we also demonstrate the slight impact of the current LORS method on inference speed, which might be improved by reducing the serial and redundant computations within LORS but not the focus of this paper.

Table~\ref{table_main2} showcases the remarkable performance of the AdaMixer + LORS method under the 3× training scheme with different backbones and query numbers. 
It can be observed that the proposed method consistently outperforms the vanilla AdaMixer across all backbones, query numbers, and evaluation metrics. This result is somewhat surprising to us, as the LORS enables the model to use significantly fewer parameters during both training and inference.

Specifically, when adopting the ResNet-50 backbone and 100 queries, LORS improves the AP value of AdaMixer by 0.5 points (43.7 vs 43.2). It is worth noting that in Table \ref{table_main1}, for experiments with the same configuration except for training 12 epochs, the AP with LORS is 0.1 lower than without LORS. However, this situation is reversed when training for 36 epochs.
The longer training time under the 3× training scheme allows LORS to be fully trained and perform better. This emphasizes the importance of sufficient training time for the LORS technique to fully exploit its potential in enhancing the performance.
Furthermore, when using more powerful backbones and a larger number of queries, the performance metrics of models utilizing LORS consistently and comprehensively surpass those of their counterparts without LORS. Even when employing a strong backbone such as Swin-S~\cite{liu2021swin} and using 300 queries, LORS still manages to further improve the model's performance while reducing neary half of the overall parameters. Much fewer parameters, yet improved performance.

In conclusion, the results in Table~\ref{table_main1} and Table~\ref{table_main2} show the effectiveness of the LORS method in reducing the number of parameters while achieving even better performance. The method's adaptability to various backbones and query numbers further demonstrates its versatility and potential for broader applications. As for why LORS can achieve better performance with fewer parameters, a plausible explanation is that after the model has been sufficiently trained and stabilized, the parameter matrix within the stacked structure is inherently sparse and low-rank. Without using LORS, the model needs to learn such a low-rank structure through training. In contrast, LORS directly and explicitly makes the structure sparse, which, to some extent, can be considered as adding a form of regularization to the model training, thereby facilitating the model's learning. 

\subsection{Ablation Study}
To save computational resources, all ablation studies are conducted using a ResNet-50 backbone and a 1× training scheme.

\begin{table}[ht]
\setlength{\abovecaptionskip}{10pt}
\setlength{\belowcaptionskip}{5pt}
\centering
\resizebox{0.66\columnwidth}{!}{
\begin{tabular}{cccc}
\toprule
LORS\textsuperscript{A} & LORS\textsuperscript{T} & \begin{tabular}{@{}c@{}}Decoder\\ Params(M)\end{tabular} & $AP$  \\
\midrule
 &   & 110 & 42.5† \\
\checkmark &   & 79 & 42.6 \\
  & \checkmark & 66 & 42.6  \\
\rowcolor{gray!20} \checkmark& \checkmark  & 35 &  42.6  \\
\bottomrule
\end{tabular}
}
\caption{Effect of LORS\textsuperscript{A} and LORS\textsuperscript{T}. "†" denotes this AP result was reproduced by ourselves.
Both LORS\textsuperscript{T} and LORS\textsuperscript{A} can reduce parameters without compromising performance.}
\label{table_whtherAN}
\end{table}

\noindent\textbf{Adaptive \& Static LORS.}
We first conduct ablation studies on the impact of 
LORS\textsuperscript{A} and LORS\textsuperscript{T} on model parameters and performance, as shown in Table \ref{table_whtherAN}. In this table, a $\checkmark$ indicates that we use the LORS method to replace the corresponding component in the AdaMixer model mentioned earlier, while a blank cell means we use the original AdaMixer module. If neither of the two is present, the model is identical to the original AdaMixer. As can be seen from the table, both adaptive and static LORS can effectively reduce the parameter size of the decoder in the model (31M and 44M, respectively) without affecting the model performance. Note that the symbol † here represents that the 42.5 AP value is reproduced by us, which is slightly different from the 42.7 AP recorded in the original paper. We believe this discrepancy can be attributed to the randomness of 1x training and does not affect our conclusions.

\noindent\textbf{Shared and private weights.}
The second point of interest is to determine which of the shared weight and private weight has a greater impact on the decoder's performance. Experiment in Table \ref{table_shared_private} is designed to address this question, which is performed on ACM and ASM of AdaMixer's decoder. Recall that the formula we use to generate the final weight parameters is $W = W^{\text{shared}} + W^{\text{private}}$. In the table, a $\checkmark$ indicates that the corresponding term is present on the right side of the formula, otherwise, it is removed. It can be seen that the absence of either term will cause a nonnegligible decrease in model performance, and the impact of missing $W^{\text{private}}$ is even more severe. This suggests that although the number of parameters in each $W^{\text{private}}$ is smaller than that of $W^{\text{shared}}$, it could be more important, possibly because the most informative part of each layer in the stacked structure is contained within it.

\begin{table}[htb]
\setlength{\abovecaptionskip}{10pt}
\setlength{\belowcaptionskip}{5pt}
\centering
\resizebox{0.66\columnwidth}{!}{
\begin{tabular}{cccc}
\toprule
$W^{\text{shared}}$ & $W^{\text{private}}$ & \begin{tabular}{@{}c@{}}Decoder\\ Params(M)\end{tabular} & $AP$  \\
\midrule
\checkmark & & 58 &  40.9\\
& \checkmark & 43 &  41.4\\
\rowcolor{gray!20} \checkmark & \checkmark & 35 & 42.6 \\
\bottomrule
\end{tabular}
}
\caption{Effect of shared parameters $W^{\text{shared}}$ and private parameters $W^{\text{private}}$ 
Default choice for our model is colored gray.}
\label{table_shared_private}
\end{table}

\noindent\textbf{Hyperparameters of $\text{LORS}^A$.}
Another aspect we would like to explore is the optimal number of parameter groups and the value of rank $r$ for adaptive LORS. Here, we feel it is necessary to explain why we use grouped parameters. Recalling formula \ref{formula:adpt_private}, the adaptive private parameters are obtained by continuous multiplication of three matrices, so the matrix with the smallest rank among them determines the rank of multiplication, which is the upper limit for the private parameters. If the task-specific requirement for the rank of the private parameters matrix unfortunately exceeds the rank of the smallest component, there will be a bottleneck when training the model. Therefore, we designed this parameter group mechanism, using the sum of the results of multiple matrix multiplication to break the aforementioned problem. Another advantage is that the rank of the final matrix possibly increases linearly with the number of parameters, while simply using a larger $r$ would lead to a quadratic increase in the number of parameters. In the end, looking at Table \ref{table_hyper_A}, we observed that the experimental result for the 6-layer decoder of the AdaMixer object detector is relatively better when using [1,1,2,2,3,3] groups of parameters per layer and $r=16$ for each group. This slightly strange parameter group numbers is inspired by AdaLoRA~\cite{zhang2023adaptive}. It suggests that higher-level neural network parameters may need to allocate more ranks when performing LoRA fine-tuning, which we suppose may also hold true for LORS, and the experiment results confirmed that.

\begin{table}[htb]
\setlength{\abovecaptionskip}{10pt}
\setlength{\belowcaptionskip}{5pt}
\centering
\resizebox{0.68\columnwidth}{!}{
\begin{tabular}{cccc}
\toprule
\begin{tabular}{@{}c@{}}Parameter\\ Groups\end{tabular} & Rank $r$ & \begin{tabular}{@{}c@{}}Decoder\\ Params(M)\end{tabular} & $AP$  \\
\midrule
1 1 1 1 1 1 & 8  & 32 &  41.8\\
1 1 1 1 1 1 & 16 & 34 &  42.1\\
2 2 2 2 2 2 & 16 & 35 &  42.4\\
\rowcolor{gray!20} 1 1 2 2 3 3 & 16 & 35 &  42.6\\
1 1 1 1 1 1 & 32 & 38 &  41.2\\
\bottomrule
\end{tabular}
}
\caption{Validation accuracy with different groups of parameters and rank $r$ in LORS\textsuperscript{A}. Default choice for our model is colored gray. Each row's first six numbers indicate the number of parameter groups sequentially used in each of the six decoder layers.}
\label{table_hyper_A}
\end{table}

\noindent\textbf{Hyperparameters of $\text{LORS}^T$.}
We further investigate the optimal configuration settings for $\text{LORS}^T$ to achieve satisfactory performance, as shown in Table \ref{table_hyper_N}. We observed that the best option is to use one group of low-rank parameters with $r=8$ for the decomposition of each layer $L_\text{output}$ in the stacked decoder of AdaMixer, which can achieve a model performance comparable to the original detector. Since using one group of parameters already yields reasonably good results, we did not explore the use of more parameter groups. Moreover, we should also note that the value of rank $r$ is not necessarily better when larger or smaller, which may indicate that the low-rank structure required by the model for a specific task has its own particularity.

\begin{table}[htb]
\setlength{\abovecaptionskip}{10pt}
\setlength{\belowcaptionskip}{5pt}
\centering
\resizebox{0.7\columnwidth}{!}{
\begin{tabular}{cccccc}
\toprule
\begin{tabular}{@{}c@{}}Parameter\\ Groups\end{tabular} & Rank $r$ & \begin{tabular}{@{}c@{}}Decoder\\ Params(M)\end{tabular} & $AP$ \\
\midrule
1 1 1 1 1 1 & 2	& 34 &  42.3\\
1 1 1 1 1 1 & 4	& 34 &  41.9\\
\rowcolor{gray!20} 1 1 1 1 1 1 & 8	& 35 & 42.6\\
1 1 1 1 1 1 & 16 & 37 & 42.1\\
1 1 1 1 1 1 & 32 & 40 & 41.7\\
\bottomrule
\end{tabular}
}
\caption{Validation accuracy with different rank $r$ in LORS\textsuperscript{T}. Default choice for our model is colored gray. Each row's first six numbers indicate the number of parameter groups sequentially used in each of the six decoder layers.}
\label{table_hyper_N}
\end{table}

\noindent\textbf{Number of decoders.}
Finally, we conclude our ablation studies with experiments on the optimal number of decoder layers incorporating the LORS structure for better performance. Since our method significantly reduces the total number of model parameters and allows each decoder layer to train and use shared parameters that capture common features, it is natural to consider whether the model's inherent characteristics are affected and whether using more or fewer decoding layers would further improve the model's performance. With this idea in mind, we conducted several experiments as shown in Table \ref{table_num_stage} to explore this issue. However, similar to the original AdaMixer without LORS, 6 layers still yield the best performance for the AdaMixer with LORS, having too many or too few layers would degrade the model's performance. This may also suggest that LORS merely captures the intrinsic structure of the model without altering its original properties.

\begin{table}[htb]
\setlength{\abovecaptionskip}{10pt}
\setlength{\belowcaptionskip}{5pt}
\centering
\resizebox{0.52\columnwidth}{!}{
\begin{tabular}{ccc}
\toprule
\begin{tabular}{@{}c@{}}Decoder\\ Number\end{tabular} & \begin{tabular}{@{}c@{}}Decoder\\ Params(M)\end{tabular} & $AP$  \\
\midrule
3   & 28 &  38.9\\
\rowcolor{gray!20} 6  & 35 &  42.6\\
9   & 42 &  42.0\\
12  & 50 &  41.0\\
\bottomrule
\end{tabular}
}
\caption{Model performance using different numbers of stacked decoders when employing LORS. The default choice for our model is colored gray. This experiment with varying numbers of decoders demonstrates that six layers still yield the best performance for the model, which is not changed by LORS.}
\label{table_num_stage}
\end{table}
\section{Conclusion}
\label{sec:5_conclusion}

This research introduced a novel approach to the reduction of parameters within deep learning models utilizing stacked structures, \ie, the Low-rank Residual Structure (LORS). It is potentially an effective methodology in reducing both static and adaptive parameters.  
Broadly speaking, LORS allows the parameters of stacked modules to be largely shared, while maintaining only a small number of unique parameters for each module, thereby significantly reducing the total number of parameters whilst not compromising performance. We validated our method via object detection task across various extensive experimental cases on the widely-used COCO dataset, and the results demonstrates a superior model performance even with a 50\% -- 70\% reduction in decoder parameters, surpassing our expectations.

There are some noticeable limitations in our study. While effective in reducing model parameters without compromising performance, enhancing model capabilities requires a comparatively long training process for the method. Additionally, our method slightly increases the inference time, as even without a detailed analysis, it can be roughly seen that LORS has unparalleled and repetitive computations needing more optimization. Finally, we have only tested our approach on object detection tasks, specifically with the AdaMixer model and its decoder structure. It's clear that our method can be applied to more tasks (e.g., NLP), different models (e.g., language models), and other neural network components (e.g., backbones). These aspects will be the focus of our future work.

\clearpage
\newpage
\section*{Appendix}
\label{sec:appendix}

\subsection*{A.1. Additional Experiments on LORS}

We would like to showcase the potential of LORS through further experiments, it shows its effectiveness in more tasks like image classification, different modules such as encoders, and across all weights of Transformers.
In fact, We managed to achieve all above goals simultaneously: we applied $\text{LORS}^{\text{T}}$ to Transformers~\cite{vaswani2017attention} within a vision encoder and used it for the classification task on CIFAR-100~\cite{krizhevsky2009learning}.

\begin{table}[ht]
\centering
\resizebox{1.0\columnwidth}{!}{
\begin{tabular}{ccccc}
\toprule
 \begin{tabular}{@{}c@{}}Attention \\ using $\text{LORS}^{\text{T}}$ \end{tabular} & \begin{tabular}{@{}c@{}} FFN \\ using $\text{LORS}^{\text{T}}$ \end{tabular} & \begin{tabular}{@{}c@{}}Parameters \\ each layer \end{tabular} & Top-1(\%) & Top-5(\%) \\
\midrule
 &   & 100\% & 63.66 & 84.23\\
\checkmark &  & 89.7\% & 63.51 & 84.85\\
  & \checkmark & 66.2\% & 63.93  & 84.69\\
\rowcolor{gray!20} \checkmark& \checkmark  & 47.5\% & 63.97 & 85.10 \\
\bottomrule
\end{tabular}
}
\caption{Effects of $\text{LORS}^{\text{T}}$ on Transformer-based DeiT encoder.}
\label{table_transformerLORS}
\end{table}

\vspace{-10pt}

\begin{table}[ht]
\centering
\resizebox{1.0\columnwidth}{!}{
\begin{tabular}{ccccc}
\toprule
 \begin{tabular}{@{}c@{}}Attention \\ Param Groups\end{tabular} & \begin{tabular}{@{}c@{}} FFN \\ Param Groups \end{tabular} & \begin{tabular}{@{}c@{}} Rank $r$ \\ per group\end{tabular} & Top-1(\%) & Top-5(\%) \\
\midrule
$\{0_{\times 12}\}$ & $\{0_{\times 12}\}$ & 32 & 59.13 & 83.11\\
$\{1_{\times 12}\}$ & $\{0_{\times 12}\}$ & 32 & 59.98 & 83.52\\
$\{0_{\times 12}\}$ & $\{1_{\times 12}\}$ & 32 & 60.43 & 83.41\\
$\{1_{\times 12}\}$ & $\{1_{\times 12}\}$ & 32 & 62.30 & 84.33\\
$\{1_{\times 9}, 2, 4, 6\}$ &  $\{1_{\times 12}\}$ & 32 & 62.92 & 83.83\\
$\{1_{\times 12}\}$ &  $\{6, 4, 2, 1_{\times 9}\}$ & 32 & 62.94 & 84.50\\
\rowcolor{gray!20} $\{1_{\times 9}, 2, 4, 6\}$ &  $\{6, 4, 2, 1_{\times 9}\}$& 32 & 63.97 & 85.10\\
\bottomrule
\end{tabular}
}
\caption{Effect of $\text{LORS}^{\text{T}}$ on DeiT with different configurations.}
\label{table_kernel_group}
\end{table}

Specifically, DeiT-Tiny~\cite{touvron2021training}, whose encoder is comprised of 12 Transformer layers, 
is trained from scratch for 300 epochs in 1 hour on CIFAR-100 with 8 V100 GPUs.
We resized images from $32 \times 32$ to $56 \times 56$, divided them into $14 \times 14$ patches as commonly done and kept original training settings except for retaining only feasible augmentations (Mixup~\cite{zhang2017mixup}, Cutmix~\cite{yun2019cutmix}, and RandomFlip).

Table \ref{table_transformerLORS} shows the main results. When applying $\text{LORS}^{\text{T}}$ to all weights in each Tranformer, the total parameters of DeiT-Tiny's encoder are reduced to 47.5\% of the original, while achieving better accuracy.

\begin{figure}[t]
  \centering
   \includegraphics[width=1.0\linewidth]{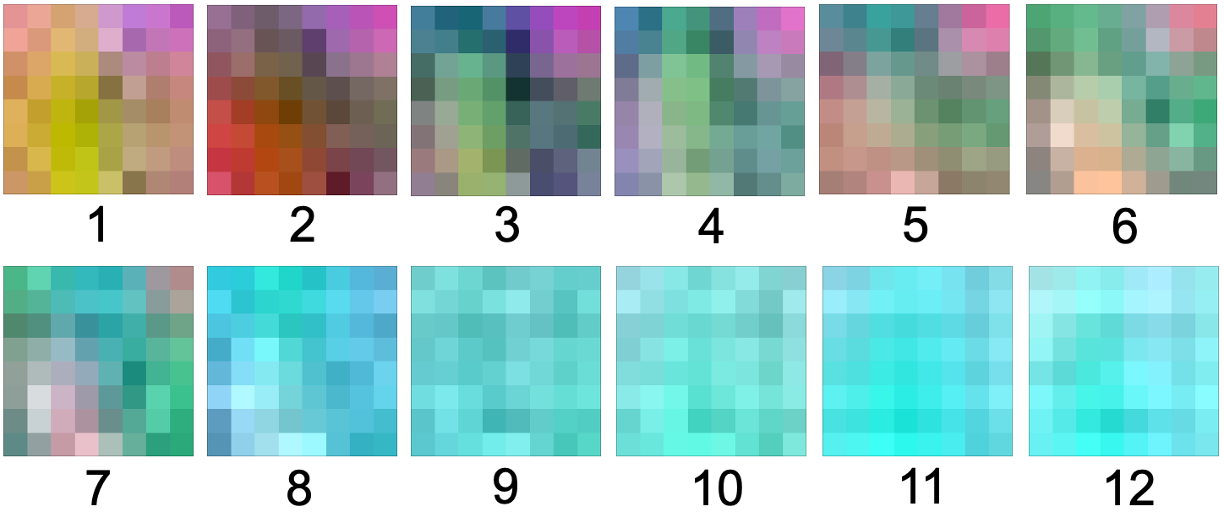}
   \caption{Visualizing input features of each layer in DeiT-Tiny. }
   \label{fig:vis_deit}
\end{figure}

Table \ref{table_kernel_group} shows an ablation study on the parameter allocation over layers. $\{1_{\times 9}, 2, 4, 6\}$ indicates that the first 9 layers of the encoder use $\text{LORS}^{\text{T}}$ with 1 group of parameters, while the last 3 layers use 2, 4, and 6 groups, respectively. The last row is our default setting, it assigns more parameters to self-attention in the last 3 layers and to FFN in the first 3 layers. 
This selection comes from the visualization of features input to each layer. We find that low-layer ones appear variable and complex, while high-layer ones appear similar and simple, as shown in Figure \ref{fig:vis_deit}. 
We hypothesize that the attention module needs more parameters to discern relationships between similar features, and FFN requires more parameters to process raw complex information. 
We select configurations for aforementioned AdaMixer experiments in a similar way. However, this empirical approach may not achieve optimal performance. 


\subsection*{A.2. Discussion of LORS with regard to RNN}

When LORS is applied to a stacked structure with only shared parameters, such a structure indeed degenerates into an RNN~\cite{rumelhart1986learning,hochreiter1997long}. However, 
The first row in Table \ref{table_shared_private} is still a hybrid recurrent architecture since it applies LORS to only part of AdaMixer~\cite{gao2022adamixer} decoders' weights, so we performed the first 4 rows in Table \ref{table_kernel_group} to facilitate this discussion. 
Its first row applied LORS to all weights in Transformer using no private parameters, fully degenerating into an RNN, and its performance is the worst. Improvement occurred weakly in the second and third rows with hybrid recurrent states, but significantly in the fourth row. 
Adding private parameters to all layers seems better than a pure RNN. These private parameters can surely be generated by a function of the previous layer, which we think could be achieved by a single $\text{LORS}^{\text{A}}$. A promising attempt might be integrating $\text{LORS}^{\text{A}}$ instead of $\text{LORS}^{\text{T}}$ into Transformers.

\subsection*{A.3. Validating the importance of self-attention and FFN in the performance of Transformers}
A natural question is whether self-attention and FFN are both crucial for Transformers' performance, as this impacts the persuasiveness of the additional LORS experiments that rely on them. Table \ref{table_wht_pretrained} shows that loading ImageNet~\cite{deng2009imagenet} pretrained weights significantly affects the performance, highlighting the importance of both components.

\begin{table}[ht]
\centering
\resizebox{\columnwidth}{!}{
\begin{tabular}{ccccc}
\toprule
 \begin{tabular}{@{}c@{}}ATTN \\ Pretrained Init \end{tabular} & \begin{tabular}{@{}c@{}}FFN \\ Pretrained Init \end{tabular} & Top-1(\%) & Top-5(\%) \\
\midrule
\checkmark & \checkmark & 78.85 & 92.73 \\
\checkmark &  & 65.28 & 86.40\\
& \checkmark & 64.19 & 84.69\\
  &  &  63.66  & 84.23\\
\bottomrule
\end{tabular}
}
\caption{Effect of whether pretrained weights loaded on attention module and feedforward module.}
\label{table_wht_pretrained}
\end{table}

{
    \small
    \bibliographystyle{ieeenat_fullname}
    \bibliography{main}
}

\end{document}